\documentclass[final]{cvpr}

\usepackage{times}
\usepackage{epsfig}
\usepackage{graphicx}
\usepackage{amsmath}
\usepackage{amssymb}

\usepackage[pagebackref=true,breaklinks=true,colorlinks,bookmarks=false]{hyperref}

\begin{document}

\title{Contextual Non-Local Alignment over Full-Scale Representation for Text-Based Person Search}

\author
{Chenyang Gao\textsuperscript{\rm 1}\thanks{This work was done when Chenyang Gao was an intern at Tencent Youtu Lab and this work was supported by 2020 Tencent Rhino-Bird Elite Training Program}, 
Guanyu Cai\textsuperscript{\rm 2},
Xinyang Jiang\textsuperscript{\rm 2}\thanks{Correspondance Author: xinyangj@zju.edu.cn},
Feng Zheng\textsuperscript{\rm 1},
Jun Zhang\textsuperscript{\rm 2}\\
Yifei Gong\textsuperscript{\rm 2}, 
Pai Peng\textsuperscript{\rm 2},
Xiaowei Guo\textsuperscript{\rm 2},
Xing Sun\textsuperscript{\rm 2}\\
\textsuperscript{\rm 1} Southern University of Science and Technology, 
\textsuperscript{\rm 2} Tencent Youtu Lab\\
}

\maketitle

\begin{abstract}
Text-based person search aims at retrieving target person in an image gallery using a descriptive sentence of that person. 
It is very challenging since modality gap makes effectively extracting discriminative features more difficult. Moreover, the inter-class variance of both pedestrian images and descriptions is small. Hence, comprehensive information is needed to align visual and textual clues across all scales. 
Most existing methods merely consider the local alignment between images and texts within a single scale (e.g. only global scale or only partial scale) or simply construct alignment at each scale separately. 
To address this problem, we propose a method that is able to adaptively align image and textual features across all scales, called NAFS (i.e. \textbf{N}on-local \textbf{A}lignment over \textbf{F}ull-\textbf{S}cale representations). 
Firstly, a novel staircase network structure is proposed to extract full-scale image features with better locality. 
Secondly, a BERT with locality-constrained attention is proposed to obtain representations of descriptions at different scales. 
Then, instead of separately aligning features at each scale, a novel contextual non-local attention mechanism is applied to simultaneously discover latent alignments across all scales. The experimental results show that our method outperforms the state-of-the-art methods by $5.53\%$ in terms of top-1 and $5.35\%$ in terms of top-5 on text-based person search dataset. \textup{The code is available at \url{https://github.com/TencentYoutuResearch/PersonReID-NAFS}}

\end{abstract}

\section{Introduction}

Text-based person search aims at retrieving target person in an image gallery using a descriptive sentence of that person. 
Compared to classical person re-identification (Re-id), it does not need an image of the target person as query which could be difficult to obtain. In addition, text-based person search is more user-friendly since it can support open-form natural language queries. Thus it has the potential to have much broader applications. 

\begin{figure}[!t]
\centering
\includegraphics[width=0.47\textwidth]{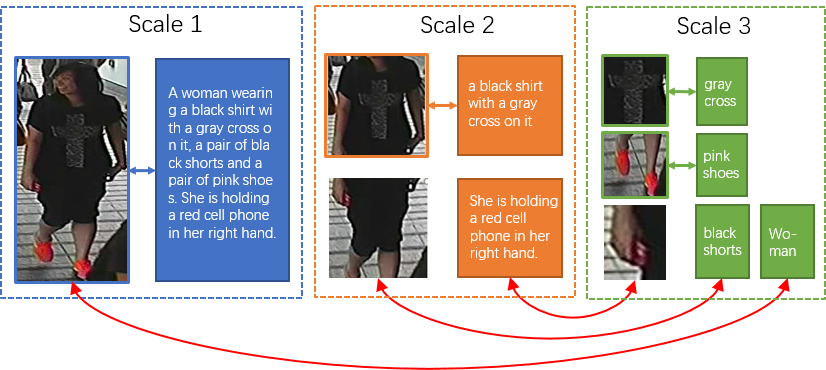} 
\caption{
Illustration of image text alignment both within similar scales and across different scales. 
} 
\label{fig_motivation}
\end{figure}

Compared with general image text matching task where an image may contain several objects, text-based person search is a much more challenging task since the high-level semantics among different pedestrian images are very similar, causing small inter-class variance of both pedestrian images and textual descriptions. 
Thus, in order to explore more distinctive and comprehensive information, text-based person search requires an algorithm to extract image and textual features from all scales. For example, both of the image and textual description in Figure \ref{fig_motivation} can be decomposed into representations at different scales. 
The sentence can be represented as short phrases, such as ``black shorts'' at scale 3, or longer sub-sentences at scale 2. 
Similarly, the image can also be partitioned into sub-regions with different sizes at scale 3 and scale 2. 
Since correct alignment between these image representations and textual representations are the basis of image text matching task, it is essential to represent the image and textual description at all scales. In this paper, we call it full-scale representation. 
However, the complex relevance at different scales makes it difficult to build a reasonable scheme of alignment. As shown in Figure \ref{fig_motivation}, in most cases, the alignment occur at similar scales, such as the sub-sentence ``a black shirt with a gray cross on it'' and the image region in scale 2, and the short phrase ``gray cross'' and the smaller image region at scale 3. 
But occasionally alignment could also occur across different scales. For instance, as shown with the red arrows in Figure \ref{fig_motivation}, a single word ``woman'' in scale 3 aligns with the whole image in scale 1. 
These phenomena illustrate the importance to jointly align image and description both within similar scales and across different scales. Therefore, a reasonable text-based person search method generally contains two key components. One is to learn image and textual representations at all scales in a coarse-to-fine fashion, the other is to explore an appropriate alignment to automatically and adaptively match these representations of different scales. 

Most of the existing works~\cite{wang2020vitaa,jing2020pose,niu2020improving}
are unable to fully satisfy the aforementioned two perspectives. 
On one hand, for multi-scale representations, most methods merely learn representations for images and textual descriptions at a certain scale. Several coarse-grained methods~\cite{li2017person,li2017identity,chen2018improving,zhang2018deep,zheng2020dual} focus on learning representations at the global scale, i.e. the whole image and sentence as shown in Figure~\ref{fig_motivation} Scale 1. Fine-grained methods \cite{wang2020vitaa,jing2020pose,niu2020improving} model the images and textual descriptions at the finest scale, e.g. image regions and short phrases as shown in Figure \ref{fig_motivation} Scale 3. 
Although some fine-grained methods \cite{jing2020pose,niu2020improving} consider combining the finest scale with the global scale, they still 
lack mid-scale information causing some description segments (image regions) fail to correctly align with proper image regions (description segments). 

On the other hand, for the cross-scale alignment, existing methods \cite{zhang2018deep,zheng2020dual,wang2020vitaa,jing2020pose,niu2020improving} try to employ pre-defined rules to align images and textual descriptions of different scales . Zhang et al. and Zheng et al.~\cite{zhang2018deep,zheng2020dual} only consider the global matching of images and textual descriptions. Some other methods~\cite{wang2020vitaa,jing2020pose} add alignments between 
short phrases and image regions as shown in Figure \ref{fig_motivation} Scale 3, but ignore alignments across different scales. Recently, Niu et al.~\cite{niu2020improving} further add extra alignments between the whole image and short phrases, as well as small image stripes and the whole sentence. These methods show that utilizing multi-scale features can significantly improve performance. However, all of them pre-define several 
alignment rules among image representations and textual representations of different scales (e.g. global-global, local-local), and build alignment within these fixed scale pairs separately. 
Hence, it limits the alignment to a certain scope, causing the alignment between image representations and textual representations outside the scale pairs are completely ignored. 

To address above problems, in this paper, we propose a novel text-based person search method that builds full-scale representations for both images and textual representations, and adaptively aligns them across all scales, called NAFS (\textbf{N}on-local \textbf{A}lignment over  \textbf{F}ull-\textbf{S}cale representations). 
First, we propose a staircase network with a novel stripe shuffling operation that incorporates better locality to the learned full scale image features. 
Then a modified BERT language model by adding a locality-constrained attention is adopted to extract full-scale textual features.
Next, instead of aligning features under several pre-defined scales (e.g., local-local, global-global), we develop a much more flexible alignment mechanism called contextual non-local attention, which is able to jointly take image representations and textual representations from all scales as input then adaptively build the alignment across all scales. 
Finally, a novel re-ranking algorithm based on the nearest visual neighbors is proposed to further improve the ranking quality. 

The main contributions of this paper can be summarized as follows: (1) A novel staircase CNN network and a local constrained BERT model are specially developed to extract full-scale image and textual representations. (2) A contextual non-local attention mechanism is proposed to adaptively align the learned representations across all scales. (3) The proposed framework achieves state-of-the-art results on the challenging dataset CUHK-PEDES\cite{li2017person}. Extensive ablation studies clearly demonstrate the effectiveness of each component in our method.

\begin{figure*}
    \centering
    \includegraphics[width=1\textwidth]{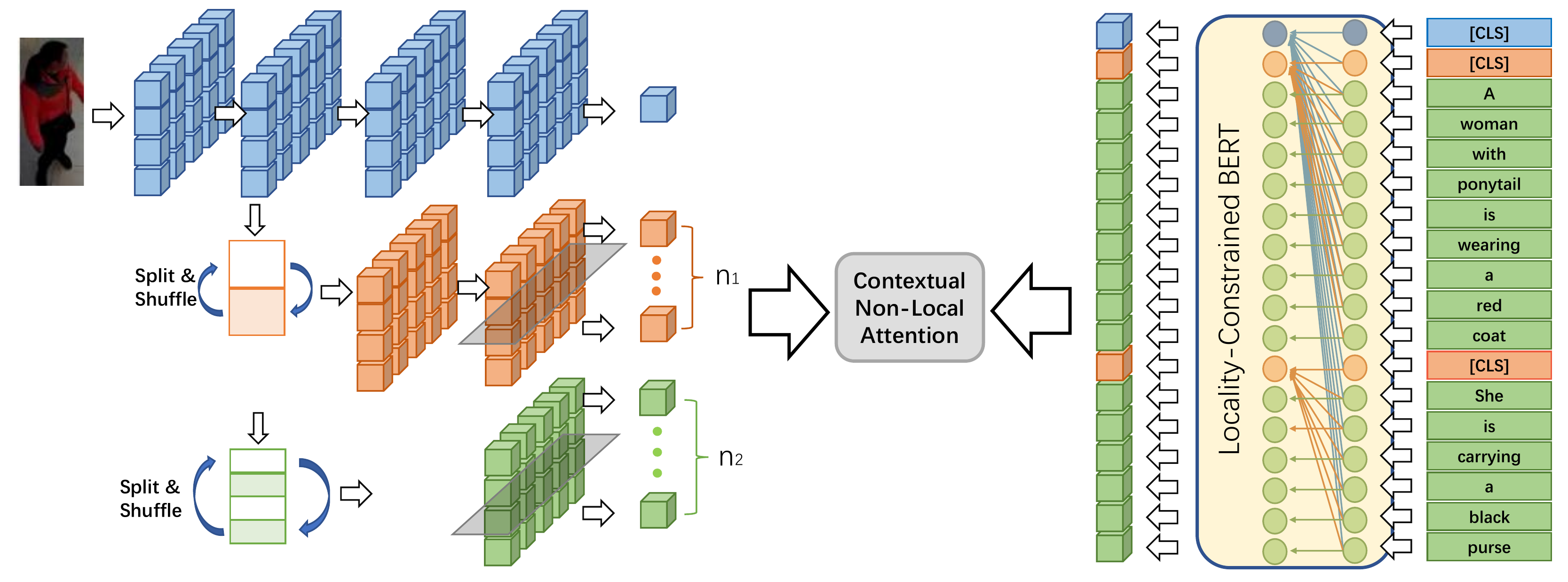}
    \caption{The overall framework consists of a stair-case network for visual representation extraction, a locality-constrained BERT for textual representation extraction and a contextual non-local attention module for join alignment. 
    }
    \label{fig_framework}
\end{figure*}

\section{Related Work}
\subsection{Person Re-identification (ReID).} 
Generally, most of the ReID methods \cite{hermans2017defense,luo2019bag} use deep CNNs to extract a global discriminative representation for each person image, while some part-based models  \cite{sun2018beyond,wang2018learning,zheng2019pyramidal} try to exploit local information. For example, PCB \cite{sun2018beyond} horizontally cuts the output feature map into six parts to learn six different local features. MGN \cite{wang2018learning} and Pyramid Network \cite{zheng2019pyramidal} propose a pyramid structured network to extract features in a coarse-to-fine manner. Moreover, some approaches propose to learn local features from local regions with semantic meanings like human part segmentation or pose \cite{zheng2019pose,liu2018pose,kalayeh2018human,su2017pose,zhao2017spindle}. However such methods highly rely on the accuracy of pose estimation and semantic parsing algorithms. 

\subsection{Text-Based Person Search.} Li et al. \cite{li2017person} first introduce the text-based person search task and propose a GNA-RNN model to learn an affinity score between the query description and the image in the gallery. 
Later, Li et al. \cite{li2017identity} propose an identity-aware two-stage network to efficiently locate simple incorrect matchings and make the result insensitive to changes in sentence structure. In \cite{chen2018improving}, a patch-wise word matching model is introduced to exploit the local matching information and obtain the proper affinity between the text and image. Zhang et al. \cite{zhang2018deep} design cross-modal objective functions for learning discriminative image-text embeddings. Moreover, a new method CMAAM \cite{aggarwal2020text} treats the task as a multi-task training framework that significantly boosts the performance of global features by introducing extra attribute annotation and prediction. Recently, Niu et al. \cite{niu2020improving} propose to define three types of alignment, namely global-global, global-local and local-local, and learn separate alignment within these three scale pairs. In addition, Wang et al. \cite{wang2020vitaa} exploit an extra segmentation model to align person partial features and textual attribute features with a k-reciprocal sampling align loss. While, a pose-guided multi-granularity attention network is explored in \cite{jing2020pose}, which contains a fine-grained alignment component and a coarse alignment component to exploit multi-granularity cross-modal relations.

\subsection{Image Text Matching.}
The goal of general image text matching is to learn a joint latent space where the embeddings of visual inputs and textual annotations can be compared directly. 
Besides global representations, some state-of-the-art methods including  SCAN \cite{lee2018stacked} and BFAN \cite{liu2019focus} also exploit alignment between images and textual fragments such as objects and words. Recently, methods like Unicoder and OSCAR \cite{li2020unicoder,li2020oscar} propose to use BERT \cite{devlin2018bert} or transformer \cite{vaswani2017attention} like network to model the text-image matching problem as a binary classification task, improving the retrieval performance at the cost of much longer inference time. 

\section{Our Method} 
In this section, we explain the proposed NAFS in detail. First, we introduce the procedures of extracting the visual and textual representations. Then we describe our contextual non-local attention mechanism. Finally, we introduce proposed re-ranking by visual neighbors to further improve the performance.
\subsection{Extracting Visual Representation}
\textbf{Staircase Backbone Structure.} Firstly, we elaborate on the implementation details of the proposed staircase network. As shown in Figure \ref{fig_framework}, 
it contains three branches, each of which is responsible for extracting visual features at different scales, from coarse to fine, namely the global branch (colored in blue), the region branch (colored in yellow) and the patch branch (colored in green). A general ResNet \cite{he2016deep} network is used as the backbone. 
1) The global branch is used to extract global and coarsest features. 
2) The region branch extracts finer features from large sub-regions in the image. It takes the feature map at the second stage of global branch as its input, then fed into two consecutive res-blocks to extract features at region scales. The output feature map of the region branch is then horizontally partitioned into $n_1$ stripes, each of which is further encoded as a local feature of a certain region. 
3) The patch branch extracts the finest features from small patches in the image. It takes the feature map at the third stage of global branch as its input, which is then fed into one res-block to extract features at small patch scales. Then we horizontally partition the output feature map into $n_2$  stripes to extract $n_2$ feature vectors for local patches.


\textbf{Split and Shuffle Operation.} A challenge of stripe-based ReID models is that due to the large perception field of CNN models, the stripe of feature maps in the deep layers may contain global information as well. 
Thus, to guarantee a better locality for the multi-scale image features, we introduce a novel split\&shuffle operation.  
It takes the intermediate feature map as input then equally partitions the feature map into several horizontal stripes denoted as a list $F$=\{$f_{1}, f_{2}, \cdots, f_{n}$\}, where $f_i$ is the $i$-th stripe starting from the top of the feature map.  
Then, such set of the partitioned stripes are randomly shuffled and re-concatenated along the vertical axis to form a complete feature map as the output. 
Both feature maps at stage 2 and stage 3 will be first split and shuffled before feeding into the range and patch branches, respectively. 
By shuffling the partitioned stripes randomly, it enables to break the inter-relationship between consecutive stripes so that the model can focus on the information within each stripe. 
Since our contextual non-local attention does not rely on the order of feature map fragments, it is not necessary to re-organize the partitioned stripes to the original order.


The visual representation extraction module takes a pedestrian image as input, and then a list of image features of different scales can be obtained and notated as $I$=\{$i_{p1},i_{p2}, \cdots, i_{pn}$\} where $i_{pi}$ $\in$ $\mathbb{R}^{D}$.

\subsection{Extracting Textual Representation}
Given a textual description $E$, we add a locality constrain to BERT to extract different scale representations of $E$. 
In our method, a textual description will be represented in three scales, separately. 
1) At the sentence-level, we add a special classification token ([CLS]) to the beginning of sentence $E$. The final hidden state corresponding to this token can be used as the sentence-level representation of the whole sentence $E$ in a global view. 
2) At the middle-level, we separate the sentence $E$ by commas resulting in a list of shorter sub-sentences. For every sub-sentence in the list, the [CLS] token is also attached to the beginning of the sub-sentence, whose final hidden state is used as the representation of each sub-sentence as well. 
3) At the finest word-level, the final hidden state of each word is directly used as the word-level representation.

For a common BERT-based model \cite{devlin2018bert}, the hidden variables of all tokens have the same global perception field. Each token can attend to any tokens in the entire input sentence. To provide locality to representations of sub-regions in a sentence (the [CLS] token for the sub-sentence), we propose a locality-constrained attention module to attend tokens within a certain range. 
Similar to the original BERT, given the query of a [CLS] token that corresponds to a sub-sentece, denoted as $q_{CLS}$, the locality-constrained attention is computed as follows:
\begin{equation}
\label{eq_locality_constrained_attention}
\text { Attention }(q_{CLS})= \sum_i \frac{e^{q_{CLS}k_i}}{\sum_ie^{q_{CLS}k_i\mathbf{1}(i\in U)}}v_i\mathbf{1}(i\in U), 
\end{equation}
where $k_i$ and $v_i$ denote keys and values corresponding to all tokens in a sentence, respectively. $U$ is the set of tokens within the range of this sub-sentence, and $\mathbf{1}(\cdot)$ is an indication function that returns $1$ when $i$-th token is in $U$. 

The textual representation extraction module takes a pedestrian description as input, and then a list of textual embeddings of different scales can be obtained and denoted as $T = \{t_{p1}, t_{p2}, \cdots, t_{pn}\}$ where $t_{pi} \in \mathbb{R}^{D}$.

\subsection{Contextual Non-Local Attention Mechanism}
\begin{figure}
    \centering
    \includegraphics[width=0.48\textwidth]{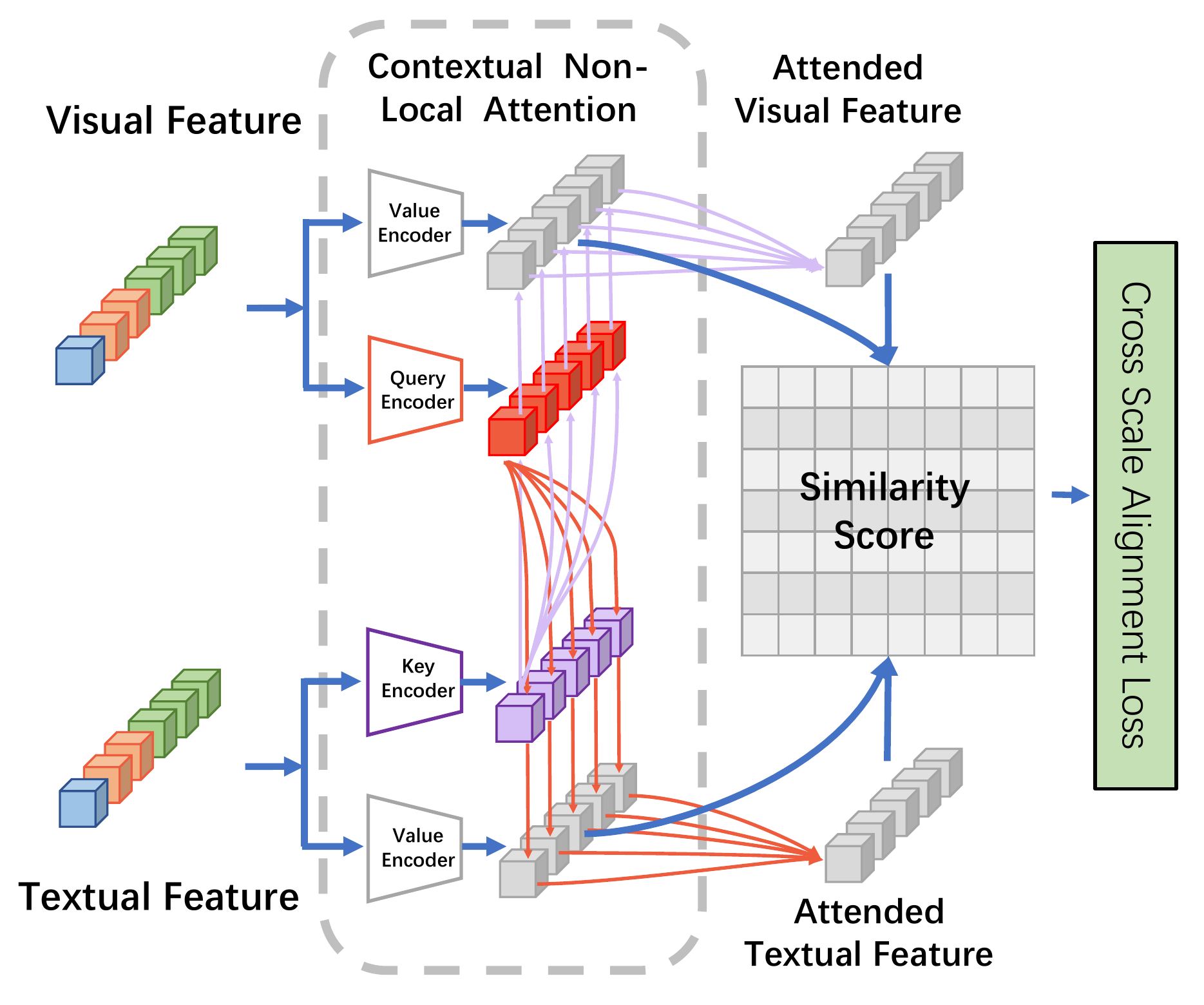}
    \caption{The illustration of proposed contextual non-local attention module. 
    }
    \label{fig_attention}
\end{figure}

As shown in Figure~\ref{fig_attention}, the contextual non-local attention expects two inputs: a set of visual features $I=\{i_{p1}, i_{p2}, \cdots, i_{pm}\}$ and a set of textual features $T=\{t_{p1}$, $t_{p2}, \cdots, t_{pn}\}$. The output of the attention module is a similarity score that measures the relevance of a image-text pair. In a nutshell, the contextual non-local attention enables cross-modal features to align with each other in a coarse-to-fine fashion according to their semantics, instead of merely using pre-defined and fixed rules (e.g., local-local, global-global).

Inspired by the spirit of self attention~\cite{vaswani2017attention}, we can explain our proposed attention mechanism as mapping a query and a set of key-value pairs to an output. 
For visual features, two learned linear projections are used to map $I$ to visual queries $I_{Q}=\{I_{q1}, I_{q2},\cdots, I_{qm}\}$ and visual values $I_{V}=\{i_{v1}, i_{v2},\cdots, i_{vm}\}$. Similarly, two linear projections are explored to map $T$ to textual keys $T_{K}=\{t_{k1}, t_{k2},\cdots, t_{kn}\}$ and textual values $T_{V}=\{t_{v1}, t_{v2},\cdots, t_{vn}\}$. 
Based on $I_{Q}$, $I_{V}$, $T_{K}$ and $T_{V}$, we introduce our proposed attention mechanism in both \texttt{Image-Text} and \texttt{Text-Image} ways.

\textbf{Image-Text Contextual Non-Local Attention.} The proposed image-Text attention module includes two stages. 
First, each visual query attend to textual keys to get a corresponding attended textual value. 
Then, considering all visual values and their attended textual values, similarity between a image-text pair can be determined. 
In detail, to obtain attended textual values, we first compute the cosine similarity matrix of $I_{Q}$ and $T_{K}$ to obtain the weights on $T_{V}$ as follows:
\begin{align}\label{eq_i2t_attn}
    s_{a,b}=[\frac{i^{T}_{qa}t_{kb}}{||i_{qa}||||t_{kb}||}]_{+},\;a\in m,b\in n,[x]_{+}=max(x,0)
\end{align}
where $s_{a,b}$ denotes the similarity between the $a$-th visual query and $b$-th textual key. Further, we normalize it as $\hat s_{a,b}=\frac{s_{a,b}}{\sum_{a=1}^{m}s_{a,b}}$. Moreover, to filter out irrelevant textual values, a focal attention trick which is similar to~\cite{Liu2019FocusYA} is used, where $\tilde{s}_{a,b}=[\sum_{c=1}^{n}\hat s_{a,b}-\hat s_{a,c}]_+\hat s_{a,b}$. Then, we compute the weighted textual values as:
\begin{align}\label{eq_i2t_weight}
    r_{va}=\sum_{b=1}^{n}\alpha_{a,b} t_{vb},\;\alpha_{a,b}=\frac{exp(\lambda_1\tilde{s}_{a,b})}{\sum_{b=1}^{n}exp(\lambda_1\tilde{s}_{a,b})}
\end{align}
where $\lambda_1$ is the inverse temperature of the softmax function.

In the second stage, we define the relevance between $a$-th visual value and its corresponding textual context using the cosine similarity between $i_{va}$ and $r_{va}$:
\begin{align}\label{eq_i2t_sim}
    R(i_{va},r_{va})=\frac{i_{va}^{T}r_{va}}{||i_{va}||||r_{va}||}    
\end{align}
By averaging all $R(i_{va},r_{va})$, we obtain the similarity of a image-text pair as
\begin{align}\label{eq_i2t_final}
    S(I,T)=\frac{\sum_{a=1}^{m}R(i_{va},r_{va})}{m}
\end{align}
As illustrated in our proposed attention mechanism, each visual feature pays more attention to relevant textual features. The relevant textual features may come from a word, a short phrase or a whole sentence, merely depending on whether the visual and textual features share similar semantics. While, instead, previous methods~\cite{wang2020vitaa,jing2020pose} tend to build the correspondence in a fixed way. We relax such constraints by enabling a semantic-based attention mechanism to build correspondence across different scales, which helps us align a image-text pair more adaptively and correctly.

\textbf{Text-Image Contextual Non-Local Attention.} Similar to the Image-Text contextual non-local attention, we regard textual keys as queries and visual queries as keys, respectively, and attend textual keys with respect to visual queries. Then, with textual values and attended visual values, we compute the similarity between a image-text pair. Specifically, the weight of $b$-th visual value with respect to $a$-th textual value is defined as $s^{\prime}_{a,b}=[\frac{t^{T}_{ka}i_{qb}}{||t_{ka}||||i_{qb}||}]_{+},\;a\in n,b\in m$. The normalized and focal attended weight is defined as $\tilde{s^{\prime}}_{a,b}=[\sum_{c=1}^{m}\hat s^{\prime}_{a,b}-\hat s^{\prime}_{a,c}]_+\hat s^{\prime}_{a,b}$, where $\hat s^{\prime}_{a,b}=\frac{s^{\prime}_{a,b}}{\sum_{a=1}^{n}s^{\prime}_{a,b}}$.

Then, we define the weighted visual value as $r^{\prime}_{va}=\sum_{b=1}^{m}\alpha^{\prime}_{a,b} i_{vb}$, where $\alpha^{\prime}_{a,b}=\frac{exp(\lambda_2\tilde{s}^{\prime}_{a,b})}{\sum_{b=1}^{m}exp(\lambda_2\tilde{s}^{\prime}_{a,b})}$. Using the weighted visual value $r^{\prime}_{va}$ and textual value, we compute the similarity of them as $R(t_{va},r^{\prime}_{va})=\frac{t_{va}^{T}r^{\prime}_{va}}{||t_{va}||||r^{\prime}_{va}||}$. The final similarity of a image-text pair is obtained by averaging operation $S^{\prime}(T,I)=\frac{\sum_{a=1}^{n}R(t_{va},r^{\prime}_{va})}{n}$.

\textbf{Alignment Objective.} We introduce an objective function named Cross-Scale Alignment Loss (CSAL) to optimize the proposed algorithm. Given a mini-batch of images $\{I_i\}^{B}_{i=1}$, captions $\{T_j\}^{B}_{j=1}$ and all image-text pairs $\{(I_i,T_j),y_{i,j}\}^{B\times B}_{i=1,j=1}$ where $y_{i,j}=1$ if $(I_i,T_j)$ is a matched pair otherwise $0$, we define the image-to-text similarity of $(I_i,T_j)$ as $S(I_i,T_j)$ and text-to-image similarity as $S^{\prime}(T_j, I_i)$. To maximize similarities between the matched pairs and restrain correspondences of unmatched pairs, we define CSAL as:
\begin{align}\label{eq_loss}
    \mathcal{L}_{CSAL}=&\mathcal{L}_{i}+\mathcal{L}_{t},\;where\\\nonumber
    \mathcal{L}_{i}=\frac{1}{B}\sum^{B}_{i=1}\sum^{B}_{j=1}S(I_i,T_j)&log\frac{S(I_i,T_j)}{q_{i,j}+\epsilon},q_{i,j}=\frac{y_{i,j}}{\sum_{k=1}^{B}y_{i,k}} \\\nonumber
    \mathcal{L}_{t}=\frac{1}{B}\sum^{B}_{j=1}\sum^{B}_{i=1}S^{\prime}(T_j, I_i)&log\frac{S^{\prime}(T_j, I_i)}{q_{i,j}+\epsilon},q_{i,j}=\frac{y_{k,j}}{\sum_{k=1}^{B}y_{k,j}}
\end{align}
where $\epsilon$ denotes a small number to avoid numerical problems.

Considering that the backbone is essential to features from multiple scales, we use Cross-Modal Projection Matching (CMPM) $\mathcal{L}_{CMPM}$ and Cross-Modal Projection Classification $\mathcal{L}_{CMPC}$ (CMPC) proposed by Zhang et al.~\cite{zhang2018deep} to stabilize the training procedure by adding CMPM and CMPC loss on features extracted from the global branch. Thus, the final objective function is:
\begin{align}\label{total_loss}
    \mathcal{L}=\lambda_{2}\mathcal{L}_{CMPM}+\lambda_{3}\mathcal{L}_{CMPC}+\lambda_{4}\mathcal{L}_{CSAL}
\end{align}



\subsection{Re-Ranking by Visual Neighbors}
We propose a multi-modal re-ranking algorithm to further improve the performance by comparing the visual neighbors of the query to the gallery (RVN). 
Given a textual query $T$, the initial ranking list is obtained by sorting the images based on their similarities to the query obtained by Eq.\ref{eq_i2t_final}. 
Then, for each image $I$ in the initial list, we obtain its $l$-nearest neighboring images based on the similarity of their visual representations, denoted as $N_{i2i}(I, l)$. 
Similarly, the nearest neighbors of the textual query can be obtained based on the similarity between its textual representations and the visual representation of images, denoted as $N_{t2i}(T, l)$. 
Here, to accelerate the computation, only the global feature is used for finding nearest neighbors. Then, we re-calculate the pair-wise similarity between the textual query and each image in the gallery by comparing the $l$-nearest neighbors with Jaccard Distance: 
\begin{align}
    D_J(I, T)=1 - \frac{N_{i2i}(I, l)\bigcap N_{t2i}(T, l)}{N_{i2i}(I, l)\bigcup N_{t2i}(T, l)}
\end{align}
Finally, the gallery is re-sorted based on the averaged scores of the original similarity and the Jaccard Distance. 

\section{Experiments}
In this section, we evaluate our proposed NAFS by comparing the person search performance with state-of-the-art methods. Furthermore, we conduct ablation studies to demonstrate the effectiveness of each component. Finally, the attentions between images and textual descriptions are visualized to demonstrate NAFS's ability to discover joint alignment across multiple scales.

\subsection{Experimental Setup}
\textbf{Dataset and Evaluation Protocol.} We evaluate our proposed model on the CUHK-PEDES dataset, which is currently the only benchmark for text-based person search.
It contains 40206 images from 13003 unique person IDs in total. 
The training set has 34054 images, 11003 person IDs and 68126 textual descriptions. 
The validation set has 3078 images, 1000 person IDs and 6158 textual descriptions. 
The test set has 3074 images, 1000 person IDs and 6156 textual descriptions. 
On average, each image contains 2 different textual descriptions and each textual description is generally longer than 23 words. 
The vocabulary of the dataset contains 9408 different words.
Following the standard evaluation setting, the performance is measured by top-k accuracy (K = 1, 5, 10). Specially, given a person description, if top-k images contain any person corresponding to the given description, the search is successful. Top-k accuracy is the percentage of successful searches among all searches.

\textbf{Implementation Details.} For the visual representation extraction module, we use ResNet-50 as our backbone for fair comparisons with previous methods. The region branch splits the feature map into two stripes equally and the patch branch splits the feature map into three stripes equally.  
The number of output strides of the convolution layer at the last stage of the backbone is set to 1.
The dimension $D$ of the image features at different scales is 768.  
We use horizontally flipping (50\% probability) as data augmenting. 
All images are normalized and resized to 384 $\times$ 128 before sending into the network. 
For the textual representation extraction module, we use BERT-Base-Uncased model as our backbone. 
The dimension $D$ of different scale textual features is set to 768 as well. 

We initialize ResNet-50 with the weights pre-trained on the ImageNet classification task. And we initialize the weights of BERT-Base-Uncased model with the weights pre-trained using a combination of masked language modeling objective and next sentence prediction on a large corpus including the Toronto Book Corpus and Wikipedia. 
The model is optimized with the Adam \cite{kingma2014adam} optimizer and the importance hyperparameters of each loss function $\lambda_{2}$, $\lambda_{3}$ and $\lambda_{4}$ in \ref{total_loss} are 1, 1 and 0.1 respectively.
The learning rate for the visual and textual feature extraction branch is set to 0.00011 and for the rest of the network layers is set to 0.0011. The batch size is 64.

\subsection{Comparison with State-of-The-Art Methods}
Table \ref{tab_sota_performance} demonstrates our results compared with state-of-the-art methods on CUHK-PEDES. GNA-RNN, CMCE and PWM+ATH use VGG-16\cite{simonyan2014very} as visual representation extraction backbone. 
While Dual Path, CMPM+CMPC, MIA, PMA, ViTAA and our NAFS use ResNet-50 as visual representation extraction backbone. 
Overall, our NAFS achieves the highest performance both with and without RVN. 
In Table \ref{tab_sota_performance}, it can be observed that methods utilizing global and local information achieve better performance than those merely use global information. This verifies the effectiveness of adapting representations at finer scale. 
Compared with ViTTA, which is the state-of-the-art method using both global and local features, NAFS gains 5.53\%, 5.35\% and 3.99\% performance improvement in terms of top1, top5 and top10 metrics respectively. This clearly verifies the effectiveness of introducing full-scale representation and contextual non-local attention mechanism.

\begin{table}
\caption{Comparison with state-of-the-art methods. Top-1,
top-5 and top-10 accuracies (\%) are reported. 
The best performance is bold. 
In the second column, ``g'' stands for global scale, ``g+l'' stands for global scale and local scale, ``m'' stands for full-scale representations from coarse to fine. ``RVN'' stands for our proposed re-ranking by visual neighbors.}
\small
\centering
\label{tab_sota_performance}
\begin{tabular}{c|c|ccc}
\hline
Method& Scale& Top1& Top5& Top10\\
\hline
GNA-RNN \cite{li2017person}& g& 19.05& -& 53.64\\
CMCE \cite{li2017identity}& g& 25.94& -& 60.48\\
PWM+ATH \cite{chen2018improving}& g& 27.14& 49.45& 61.02\\
Dual Path \cite{zheng2020dual}& g& 44.40& 66.26& 75.07\\
CMPM+CMPC \cite{zhang2018deep}& g& 49.37& -& 79.27\\
\hline
MIA \cite{niu2020improving}& g+l& 53.10& 75.00& 82.90\\
PMA \cite{jing2020pose}& g+l& 53.81& 73.54& 81.23\\
ViTAA \cite{wang2020vitaa}& g+l& 55.97& 75.84& 83.52\\
\hline
NAFS(ours) & m&  59.94& 79.86& 86.70\\
NAFS with RVN (ours)& m&  \textbf{61.50}& \textbf{81.19}& \textbf{87.51}\\
\hline
\end{tabular}
\end{table}

\subsection{Extensive Ablation Studies}
\begin{table}[t]
\centering
\caption{Comparison with the methods using feature representations at different scales. Top-1, top-5 and top-10 accuracies  (\%) are reported.}
\label{tab_different_level_feature_performance}
\begin{tabular}{c|ccc}
\hline
Feature& Top1& Top5& Top10\\
\hline
Global& 55.47& 77.29& 84.36\\
Local+Global& 56.90& 77.92& 84.81\\
Full Scale& 59.94& 79.86& 86.70\\
\hline
\end{tabular}
\end{table}

\begin{table}
\centering
\caption{Performance Comparison of separate alignment and joint alignment. Top-1, top-5 and top-10 accuracies (\%) are reported.}
\label{tab_different_alignment_performance}
\begin{tabular}{c|ccc}
\hline
Method& Top1& Top5& Top10\\
\hline
Separate Alignment& 57.98& 78.22& 85.43\\
Joint Alignment& 59.94& 79.86& 86.70\\
\hline
\end{tabular}
\end{table}

\begin{table*}
\centering
\caption{Performance comparison of different components in our methods. Top-1, top-5 and top-10 accuracies (\%) are  reported.}
\label{tab_different_module_performance}
\begin{tabular}{ccccc|ccc}
\hline
BERT& Staircase Network& Contextual Non-local& Split\&Shuffle & RVN & Top1& Top5& Top10\\
\hline
$\times$& $\times$& $\times$& $\times$& $\times$& 54.76& 77.10& 84.86\\
$\surd$& $\times$& $\times$& $\times$& $\times$& 55.47& 77.29& 84.36\\
$\surd$& $\surd$& $\times$& $\times$& $\times$& 57.59& 78.22& 85.71\\
$\surd$& $\surd$& $\surd$& $\times$& $\times$& 59.63& 79.53& 86.42\\
$\surd$& $\surd$& $\surd$& $\surd$& $\times$& 59.94& 79.86& 86.70\\
$\surd$& $\surd$& $\surd$& $\surd$& $\surd$& 61.50& 81.19& 87.51\\
\hline
\end{tabular}
\end{table*}

\begin{figure*}
    \centering
    \includegraphics[width=1.01\textwidth]{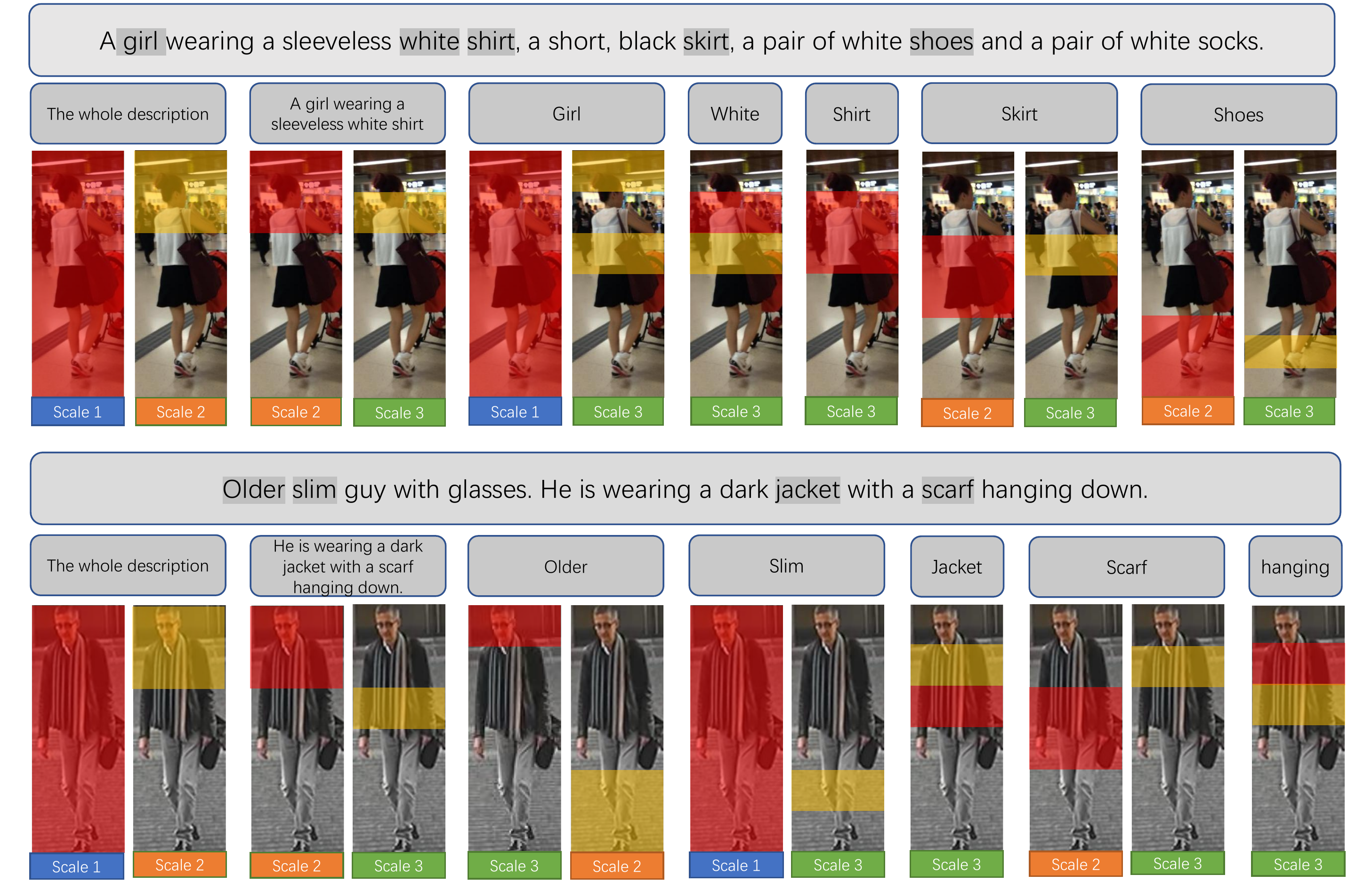}
    \caption{Visualization of the joint alignment between words and image regions across different scales. The image regions highlighted in red and yellow color are the ones have the highest and the second highest attention weights to the corresponding words.
    }
    \label{fig_visualization}
\end{figure*}

\textbf{Full-Scale Representation and Joint Alignment.} We conduct ablation studies to compare the performance of using image and textual representations at different scales. The results of the following three methods are shown in Table 2.
\begin{itemize}
    \item \textbf{Global Features.} This method only extracts global features for the entire images and descriptions. Since contextual non-local attention is not suitable for methods with only global representations, only CMPM and CMPC loss is applied to train the model. 
    \vspace{-4pt}
    \item \textbf{Local + Global Features.} Besides the global features, this method further adds the image and text representations at the finest scale (scale 3). 
    Apart from missing the mid-scale representations, the other components are exactly the same as our proposed method. We select this method to verify the effectiveness of adding finest image and textual representations to text-based person search tasks. 
    \vspace{-4pt}
    \item \textbf{Full Scale Features.} This is the full implementation of our proposed NAFS, with images and textual representations at three different scales, from coarse to fine. This comparison is to verify the effectiveness of adding representations of mid-level scale (scale 2).
\end{itemize}
Table \ref{tab_different_level_feature_performance} shows the performance of using representations under different scales. It is observed that the Top1 performance increases from 55.47 to 56.90 after adding local information. After adding mid-scale information, the top1 performance increases from 56.90 to 59.94. This implies different scale information is beneficial to the alignment procedure.

In order to verify the effectiveness of introducing joint alignment to the representations across different scales, we compare our joint alignment with methods using pre-define alignment.
As shown in  Table \ref{tab_different_alignment_performance}, separate alignment refers to separately aligning image and text within three scale pairs: the whole image to the whole sentence, large image regions (n1-stripe partition) to sub-sentences and small image regions (n2-stripe partition) to words. 
From table \ref{tab_different_alignment_performance}, we observe that our proposed joint alignment outperforms the separate alignment. This verifies that jointly aligning images and textual descriptions across different scales effectively boosts the performance of text-based person search. 

\textbf{Model Components.} We divide our proposed methods into 5 different components and observe the performance improvement by adding each component, as shown in Table \ref{tab_different_module_performance}: 
\begin{itemize}
\item \textbf{Baseline.} The first row of Table  \ref{tab_different_module_performance} is a baseline model without any NAFS components. A standard bi-LSTM\cite{hochreiter1997long} and ResNet-50 is used for feature extraction. CMPM and CMPC loss is used for training. 
\vspace{-4pt}
\item \textbf{BERT.} We replace the bi-LSTM in the baseline with our proposed locality-constrained BERT (denoted as 'BERT' in Table \ref{tab_different_module_performance}). The features from different scales are concatenated together to obtain one unified feature representation for image text matching. The locality-constrained BERT gains $0.71$ performance improvement in terms of top1 accuracy.
\vspace{-4pt}
\item \textbf{Staircase Network.} The normal ResNet-50 backbone is replaced with the proposed staircase backbone structure that extracts representations at multiple scales. The features from different scales are concatenated together to obtain one unified feature vector for image text matching. The staircase network brings $2.12$ performance improvement in terms of top1 accuracy. 
\vspace{-4pt}
\item \textbf{Contextual Non-Local.} Instead of concatenating the multi-scale features, the joint alignment by the contextual non-local attention mechanism is applied, which gives $2.04$ performance improvement in terms of top1 accuracy.  
\vspace{-4pt}
\item \textbf{Split\&shuffle.} The split and shuffle operation is added to the staircase backbone structure. 
\vspace{-4pt}
\item \textbf{RVN.}  Our proposed re-ranking method by visual neighbors is applied after the initial ranking, improving the top1 accuracy by $1.56$. 
\end{itemize}

\subsection{Visualization Analysis}
To demonstrate NAFS's ability to discover joint alignment across different scales, we visualize the alignment results between textual descriptions and image regions at different scales, which is shown in Figure \ref{fig_visualization}. For better visualization of proposed contextual non-Local attention mechanism, we horizontally partition the output feature map into three stripes in region branch and six stripes in patch branch respectively. The image regions highlighted with red and yellow colors have the highest and the second highest attention weights to the corresponding textual descriptions. In the case of two sub-regions with similar attention weights, both of them will be highlighted. 

From  Figure \ref{fig_visualization}, we observe that NAFS is able to align textual descriptions with image regions across different scales, from coarse to fine. As shown in the top half of Figure \ref{fig_visualization}, the whole description aligns with the whole image and the sub-sentence ``A girl wearing a sleeveless white shirt'' aligns with the image part at scale 2. Word ``girl'' aligns with the whole image at scale 1, because a person's gender is determined by the clues from the entire image. Words ``white'' and ``shirt'' align with the image regions at scale 3 because a small part of the images contains the white shirt. Words like ``skirt'' and ``shoes'' align with both scale 3 and scale 2 image regions because the object skirt and shoes exists in both small and mid-level image regions. 
Similarly, in the bottom half of Figure \ref{fig_visualization}, the whole description aligns with the whole image and the sub-sentence ``He is wearing a dark jacket with a scarf hanging down'' aligns with the image part at scale 2. Word ``older'' matches the top image stripe at scale 3, because we can tell this man's age by his face, while the word ``slim'' matches the whole image because we need to example the full body of the person to tell if he is slim.
The visualization results verify the effectiveness of proposed joint alignment and the necessity of full-scale representations.

\section{Conclusion}

We propose a novel text-based person search method that conducts joint alignment over full-scale representations, called NAFS.
A novel staircase CNN network and a locality-constrained BERT model are proposed to extract multi-scale image and textual representations. A contextual non-local attention mechanism adaptively aligns the learned representations across different scales. Extensive ablation studies on the CUHK-PEDES dataset demonstrate that our approach outperforms state-of-the-art methods by a large margin.

\end{document}